\pdfoutput=1

\documentclass[11pt]{article}

\usepackage[final]{coling}

\usepackage{times}
\usepackage{latexsym}

\usepackage[T1]{fontenc}

\usepackage[utf8]{inputenc}

\usepackage{microtype}

\usepackage{inconsolata}

\usepackage{graphicx}

\title{Reasoning Over the Glyphs: \\
Evaluation of LLM’s Decipherment of Rare Scripts}

\author{Yu-Fei Shih\\
  National Taiwan University \\
  \texttt{yfshih@nlg.csie.ntu.edu.tw} \\\And
  Zheng-Lin Lin \\
  National Taiwan University \\
  \texttt{b09208026@ntu.edu.tw} \\\And
  Shu-Kai Hsieh \\
  National Taiwan University \\
  \texttt{shukaihsieh@ntu.edu.tw} \\\
  }

\begin{document}
\maketitle
\begin{abstract}
We explore the capabilities of LVLMs and LLMs in deciphering rare scripts not encoded in Unicode. We introduce a novel approach to construct a multimodal dataset of linguistic puzzles involving such scripts, utilizing a tokenization method for language glyphs. Our methods include the Picture Method for LVLMs and the Description Method for LLMs, enabling these models to tackle these challenges.
We conduct experiments using prominent models, GPT-4o, Gemini, and Claude 3.5 Sonnet, on linguistic puzzles. Our findings reveal the strengths and limitations of current AI methods in linguistic decipherment, highlighting the impact of Unicode encoding on model performance and the challenges of modeling visual language tokens through descriptions. Our study advances understanding of AI's potential in linguistic decipherment and underscores the need for further research.
\end{abstract}

\section{Introduction}


In an era where AI transforms our understanding of language, deciphering rare and obscure scripts presents a unique frontier. Though still in use, these writing systems often remain enigmatic due to their complexity and limited exposure. This research delves into the capabilities of large language vision models (LVLMs) and large language models (LLMs) in deciphering such scripts. This paper aims to evaluate how effectively these models can interpret and understand rare writing systems. This study advances our knowledge of AI's potential in linguistic decipherment and sheds light on the broader implications for cross-cultural communication and the preservation of human knowledge.



\section{Related Work}
Related work might be related to AI-driven decipherment of ancient scripts~\citep{wang2024opendatasetoraclebone,guan2024decipheringoraclebonelanguage,zhang2023bridginggapdecipheringtabular} and lost languages~\citep{luo2019neuraldeciphermentminimumcostflow}, which has seen significant advancements in recent years by leveraging deep learning to tackle these complex challenges. While these efforts have shown promise, challenges remain in dealing with incomplete datasets, the lack of contextual information, and the inherent ambiguity of ancient scripts. 


Recent interest has focused on AI's ability to solve the so-called `Rosetta Stone' linguistic problems~\citep{bozhanov-derzhanski-2013-rosetta} as implemented in the Linguistic Olympiads contest~\citep{vaduguru-etal-2021-sample,lin-etal-2023-solving}. These problems present a diverse array of challenges, ranging from decoding unfamiliar scripts to identifying linguistic patterns in languages with minimal exposure. 

For instance, the problems from UKLO (United Kingdom Linguistics Olympiad)\footnote{https://www.uklo.org/}are designed to test reasoning, pattern recognition, and symbolic manipulation—skills directly applicable to the task of deciphering unknown or rarely known languages and writing systems. The Benchmarks of Linguistics Olympiad-style puzzles, which test few-shot reasoning in the context of LLMs, have been proposed~\cite {chi-etal-2024-modeling, lingoly}. By applying AI models to these structured, well-defined problems, we can systematically assess their ability to generalize across different types of linguistic challenges. This approach not only provides a controlled environment for testing but also offers valuable insights into the strengths and limitations of current AI methodologies in the broader context of linguistic decipherment and archaeology.


\section{Methods}

In this section, we outline the methods used to construct a multimodal dataset of linguistic puzzles containing writing systems that cannot be encoded in Unicode. This inherently multimodal dataset addresses aspects that previous benchmarks have overlooked. Our approach enables both LVLMs and LLMs to tackle these challenges effectively.

\begin{figure}
    \centering
    \includegraphics[width=0.7\linewidth]{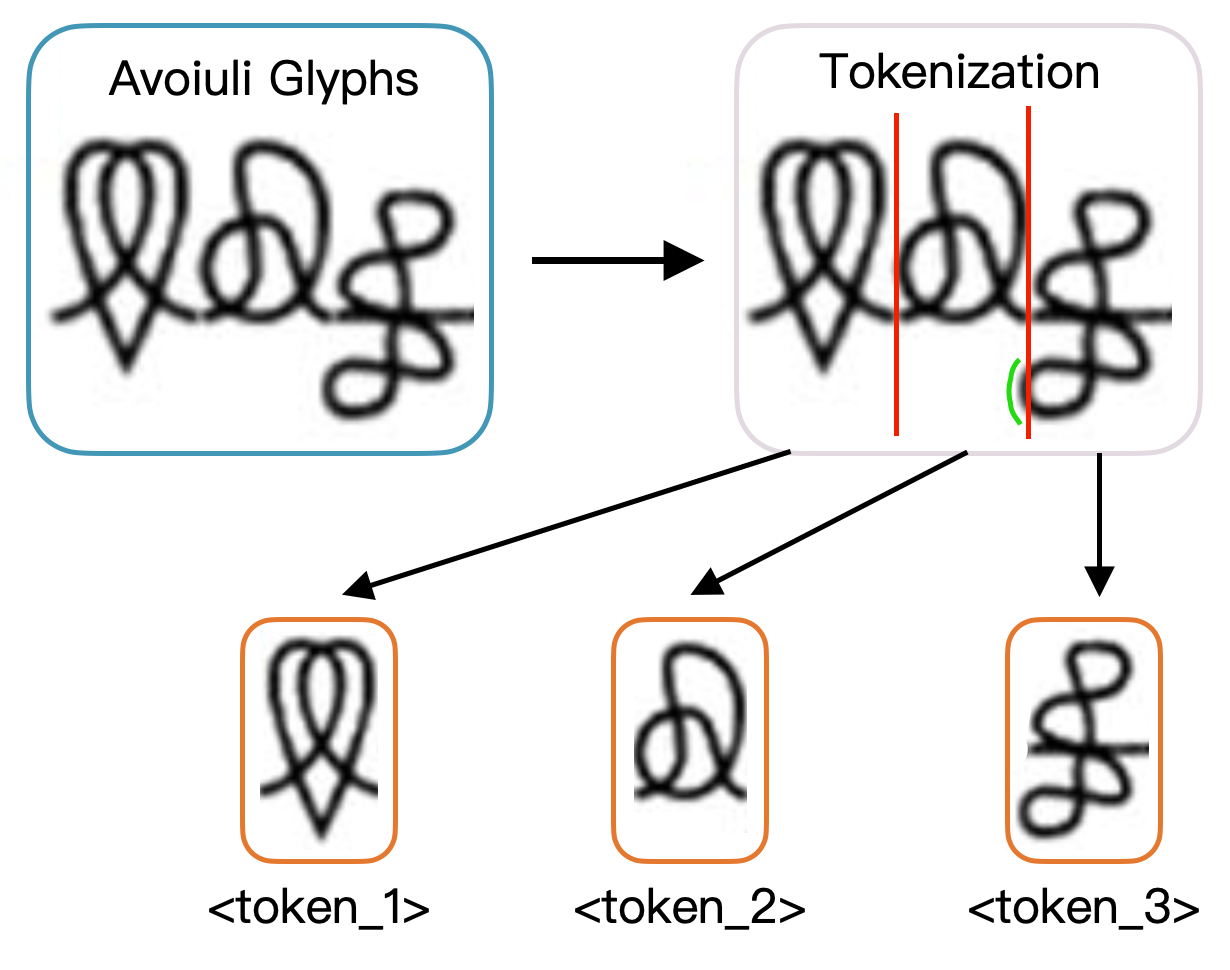}
    \caption{Example of Tokenization of Glyphs}
    \label{fig:tok}
\end{figure}

To enable models to understand scripts that are not Unicode-encodable, we introduce the concept of a \textbf{glyph token} (The tokens mentioned below all refer to glyph tokens.), representing the fundamental unit of visual information extracted from an unknown language. We propose a visual tokenization technique that segments consecutive glyphs into glyph tokens. Specifically, two adjacent glyphs are considered separate tokens if a vertical white line can pass through their gap. An exception to this rule applies to glyphs featuring horizontal extensions at the top or bottom that may bridge over the white line. Even if the white line does not intersect these glyphs due to such extensions, they are still considered separate tokens. The criteria for identifying and applying this exception may vary among individuals. Figure~\ref{fig:tok} shows the example of our tokenization on Avoiuli glyphs.

After tokenization, we replace each token with a placeholder \texttt{<token\_i>} to build the text dataset. To incorporate visual information, we propose two methods that enable LVLMs and LLMs to understand the problem.

\noindent \textbf{Picture Method}: For LVLMs, we provide an image containing all tokens with their corresponding placeholders labeled beneath them. Along with the textual script, we believe that the reasoning abilities of LVLMs will enable them to associate each placeholder in the text with the matching glyph in the image.

\noindent \textbf{Description Method}: We propose a two-stage approach for LLMs. First, we construct a description table where each token is represented by a detailed description, including its visual appearance and relational attributes to other tokens. We believe that LLMs can reference each placeholder in the problem to the corresponding description in the table.

\section{Experiment}
We conducted preliminary experiments to evaluate how LVLMs perform using our proposed methods. We selected three prominent LVLMs for our experiments: GPT-4o\citep{openai2024gpt4}, Gemini\citep{geminiteam2024geminifamilyhighlycapable}, and Claude 3.5\citep{claude3.5} Sonnet\footnote{All inputs will be made public upon acceptance.}.

\noindent \textbf{Non Unicode-Encodable Puzzles}: We chose three puzzles from the UKLO that cannot be encoded using Unicode, based on UKLO's difficulty levels. Mandombe falls under the "Advanced" category, Avoiuli appears in both "Intermediate" and "Advanced," and Ditema tsa Dinoko is categorized under "Foundation" and "Intermediate."

\noindent \textbf{Unicode-Encodable Puzzles}: To evaluate our methods on Unicode-encodable puzzles, we selected Malayalam from NACLO (North American Computational Linguistics Open competition)\footnote{https://naclo.org/} as a common language, and Meroitic from UKLO as a low-resource language.

In both experiments, the models were tasked using the Picture Method, the Description Method, and a pure text script containing only placeholders. For Unicode-Encodable puzzles, we additionally tested scripts that used the actual Unicode characters of the languages.

\noindent \textbf{Token Description Generation}: We employed models to generate a description of up to 35 words for each token, given an image containing multiple tokens. To evaluate the description generation capabilities of the models, we conducted a description pairing experiment. Each model was presented with the same image. The models then needed to associate each token with its generated description, allowing us to assess whether the descriptions effectively differentiate the tokens within the image.

\section{Result and Discussion}
\subsection{token Description Generation} 


Our experimental results showed that GPT-4o achieved an average accuracy of 40.0\%, Gemini achieved 13.4\%, and Claude-3.5 achieved 31.3\%. These results should be interpreted with caution. Models producing lower-quality descriptions often performed poorly in the description pairing task, leading to unstable accuracy.
Manual inspection of the generated descriptions revealed that none of the models could produce descriptions that accurately captured the features of complex tokens. Only tokens resembling basic geometric figures were well described by all LVLMs. The following three common mistakes were observed across all models:

\noindent \textbf{Lack of Details or Wrong Details}: A desired description should clearly mention the components and their relative positions within a token. Without correct details, it is difficult to reconstruct the token's appearance based on the description alone.

\noindent \textbf{Direction Confusion}: LVLMs often lacked a clear understanding of the directions (top, bottom, left, right) in certain tokens, frequently confusing these orientations in their descriptions.

\noindent \textbf{Misleading Elaboration}: Sometimes, models generated misleading elaborations that undermined an otherwise well-structured token description.

The token description examples are presented in Appendix~\ref{sec: table description}. The observed limitations in the LVLMs' ability to generate accurate and detailed token descriptions highlight the challenges in modeling the visual structure of language tokens through narrative descriptions. Improving the descriptive capabilities of LVLMs for intricate visual tokens remains an open area for further research.

\begin{figure}
    \centering
    \includegraphics[width=\linewidth]{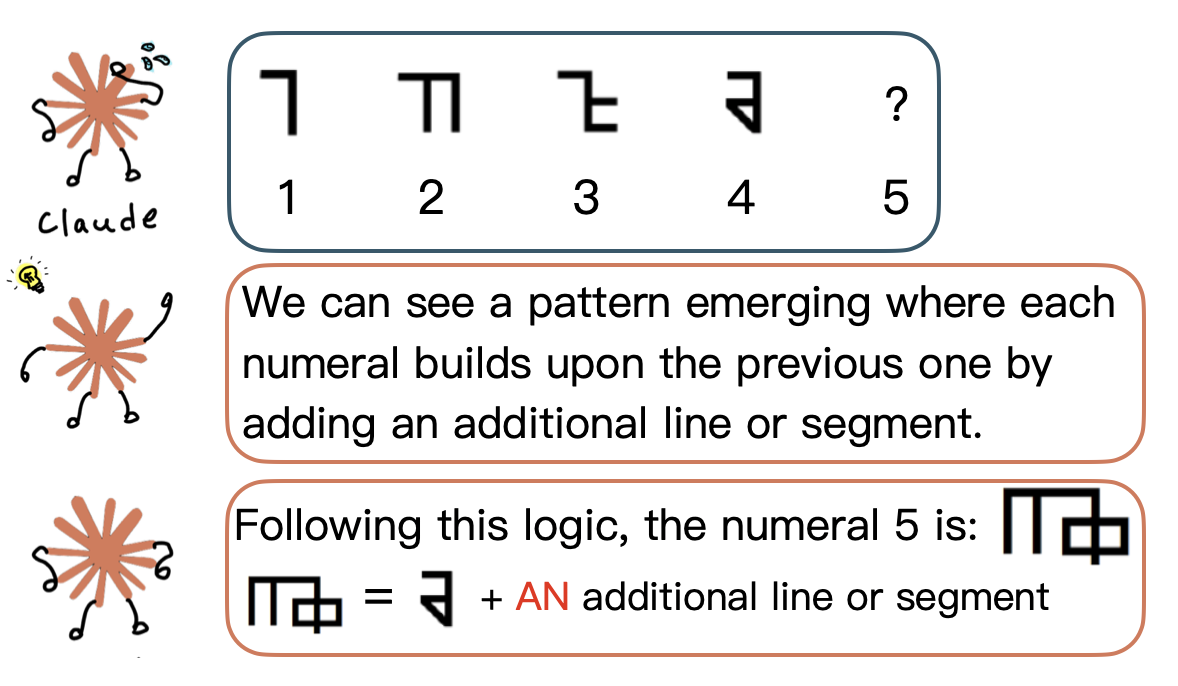}
    \caption{Incorrect geometric interpretations example.}
    \label{wrong_reasoning}
\end{figure}

\subsection{Non Unicode-Encodable Problems}

By comparing the responses of three LVLMs across different tasks and puzzles, we observed that supplying only tokens placeholder, without images or descriptions, resulted in minimal reasoning from the LVLMs.
Their responses were often limited to simple pattern matching, such as identifying token occurrences in different words and guessing the corresponding syllables, but these guesses frequently lacked a logical chain of reasoning and were often speculative.
When token descriptions were provided, the LVLMs focused on matching shared features and used these similarities for inference; although conclusions were not always correct, the reasoning was more comprehensible. Conversely, when token images were provided, the LVLMs often produced incorrect geometric interpretations and flawed reasoning, as illustrated in Figure~\ref{wrong_reasoning}, where Claude erroneously inferred numbers from token glyphs. These issues consistently arose in puzzles like Mandombe, where glyphs contain smaller components that affect syllable pronunciation. For Mandombe, the success rate of correctly reverse-mapping token images from descriptions was below $15\%$ across all three LVLMs, suggesting that providing images sometimes may even hinder reasoning abilities. In summary, the LVLMs employed several strategies across the three puzzles and input types: identifying recurring tokens and inferring corresponding syllables; comparing token descriptions to find shared features and infer relationships between tokens; analyzing geometric features in descriptions to guess meanings; and attempting geometric reasoning based on token glyph images, despite frequent errors.

Solving these three puzzles relies on two distinct approaches. For puzzles where tokens are indivisible into smaller components, such as Avoiuli, the primary challenge is identifying each token, associating it with the corresponding syllable, and constructing a syllable-token mapping table. Conceptually, even without token descriptions or glyph images, accurate inferences can be made by comparing word lengths and complexities, such as the number of different tokens required to form a word. Experimental results indicate that all three LVLMs followed a correct reasoning direction, and providing images further encouraged inference. However, the LVLMs were unable to identify all syllables represented by the tokens fully. Additionally, Avoiuli glyphs have the unique property that words remain the same when mirrored horizontally; thus, mirrored glyphs should be considered the same token. Although this meta-knowledge was not provided to solvers, and we treated mirrored glyphs as distinct tokens during tokenization, it is expected that LVLMs might infer this from images or descriptions. Nonetheless, none of the models made such inferences during the experiment.

For puzzles where the tokens contain smaller components that can alter syllable pronunciation, such as Mandombe and Ditema tsa Dinoko, the key lies in decoding each component's meaning through detailed descriptions or corresponding token images. As previously mentioned, these scripts tend to be more complex, and none of the observed model's strategies targeted phonological analysis. As a result, when faced with these types of puzzles, the LVLMs did not progress in the correct direction. Occasionally, correct answers were guessed due to the nature of the questions, such as matching or multiple-choice, but overall performance remained suboptimal.

\subsection{Impact of Unicode Encoding}

We investigate how Unicode encoding affects the performance of language models on linguistic puzzles involving both common and low-resource languages. By selecting problems that do not require token visualization, we isolate the effect of Unicode representation.

For the common Malayalam, models exhibited strong problem understanding and successfully applied grammatical rules when Unicode encoding was used, generating mostly correct answers. Notably, Gemini sometimes relied on its own grammatical rules rather than the provided ones, indicating the influence of pre-trained knowledge of Malayalam. However, when Unicode encoding was replaced with our tokenization method using placeholders \texttt{<token\_i>}, the models' abilities to apply grammar rules and understand the problem diminished significantly. GPT-4o failed to utilize grammatical clues for vocabulary translation. Gemini showed limited chain-of-thought capabilities and often misinterpreted problem requirements despite handling grammatical processing reasonably well. Claude 3.5 performed unexpectedly well on our Picture Method, possibly leveraging pre-trained Malayalam knowledge by associating placeholders with known tokens, but its performance declined without visual support. In contrast, all models faced significant challenges for the low-resource language Meroitic regardless of Unicode encoding. They failed to deduce key linguistic rules such as right-to-left writing and inherent vowels. The outputs were often contradictory and incoherent, likely due to the scarcity of training data. Even with correct Unicode tokenization, which provides directional information, performance did not improve, suggesting that Unicode support alone is insufficient without adequate training data.

These results demonstrate that Unicode encoding allows models to tap into pre-trained knowledge for common languages but has little effect on low-resource languages. Our tokenization method is a valuable alternative to assess models' true reasoning and problem-understanding abilities without the influence of pre-trained knowledge.

\subsection{Limitation and future work}
Our analysis is limited to five linguistic puzzles, which may not capture the general patterns of LVLM behaviors across a broader range of linguistic challenges. Future studies should test these models on a wider array of linguistic puzzles to generalize the findings. Additionally, we used the same prompt for each model to evaluate their behavior. In practical problem-solving scenarios, prompts should be tailored to suit each LVLM and the specific type of language puzzle.

Future work includes incorporating more Unicode and non-Unicode-encodable problems to create the dataset for further research. We also aim to apply a wider variety of prompting strategies, including few-shot in-context learning, to assess model performance across different metrics.

\section{Conclusion}
In this study, we have investigated the capabilities of LVLMs and LLMs in deciphering rare scripts that cannot be encoded in Unicode. By introducing a multimodal dataset specifically designed for linguistic puzzles involving rare scripts and developing a tokenization method tailored to non-Unicode-encoded scripts, we have provided a new framework for models to tackle these unique linguistic tasks. Our comparative performance analysis across GPT-4o, Gemini, and Claude 3.5 Sonnet highlights the strengths and limitations of each model under various input formats.

Our experiments demonstrate that current models face significant challenges when handling non-Unicode-encodable scripts or low-resource languages, including difficulties in token description generation and geometric reasoning. These limitations underscore the complexity of linguistic decipherment tasks and the need for more advanced modeling techniques.

The impact of Unicode encoding on model performance further underscores how pre-trained knowledge influences models' reasoning abilities. Our findings suggest that Unicode support alone is insufficient for low-resource languages, pointing to the necessity for models to develop genuine reasoning and problem-solving skills independent of prior knowledge. Future work should focus on expanding the dataset with more diverse puzzles, exploring varied prompting strategies, and enhancing descriptive capabilities. Addressing these challenges will advance our ability to use AI to preserve human knowledge by deciphering rare and ancient scripts, underscoring the broader impact and novelty of our approach.

\section{Appendices}
\subsection{Table description Example}\label{sec: table description}
\begin{figure}
    \centering
    \includegraphics[width=\linewidth]{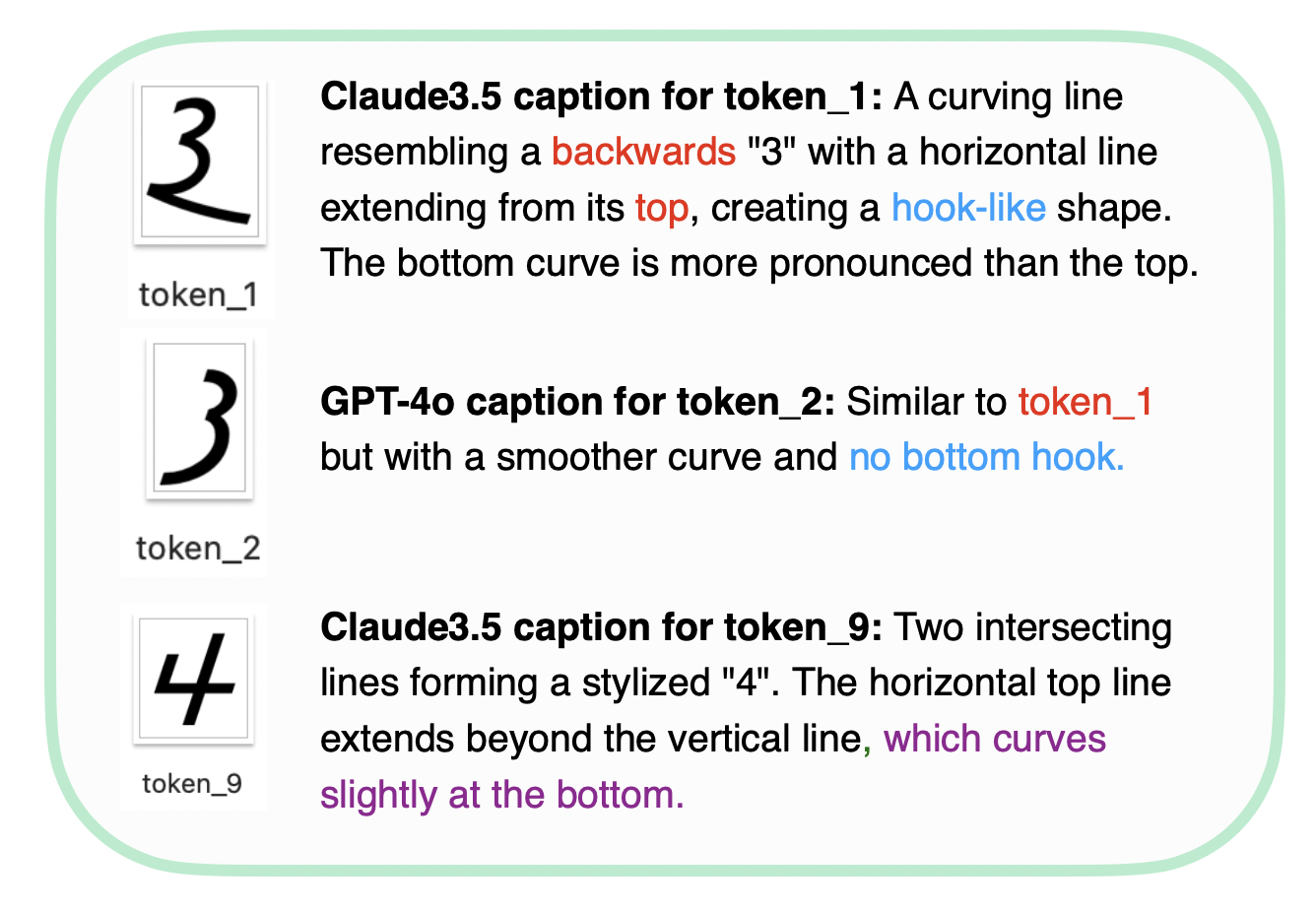}
    \caption{Example for Token's Description on Meroitic}
    \label{fig:token_description_example}
\end{figure}

Figure~\ref{fig:token_description_example} illustrates these mistakes and the desired effect we expect from the models. In the \texttt{token\_2} example, the description compares two tokens, clearly specifying the components, their relationships within the token, and their relationship with other tokens. This represents the desired behavior we expect when the models are given an image with multiple tokens.
\subsection{Analysis of Language Model Performance on Unicode-encodable Linguistic Puzzles}\label{sec: detail analysis}
This appendix presents an analysis of the performance of GPT-4o, Gemini, and Claude 3.5 on linguistic puzzles involving both common and low-resource languages. Methods including the Picture Method, Description Method, token placeholders only, and Unicode script versions were employed to assess the models' reasoning abilities and reliance on pre-trained knowledge.

\subsubsection{Performance on Malayalam Puzzles}
The Malayalam puzzles comprised three tasks: translation from Malayalam to English, grammatical tree construction, and incorrect grammar detection with explanation.

\noindent \textbf{Without Unicode Encoding}
When presented with token placeholders instead of Unicode encoding, GPT-4o often failed to apply the provided grammatical rules effectively. The model tended to translate problems directly without considering grammatical order and lacked integration between the provided translations and grammatical rules to deduce other translations. Its reasoning in constructing grammatical trees was generally incorrect.

Gemini struggled to comprehend the problem requirements. When instructed to construct grammatical trees using sentences containing verb modifiers (\textit{v-mod}), it often selected sentences arbitrarily. Despite these shortcomings, Gemini's grammatical processing was relatively adequate. The presence or absence of visual aids did not significantly affect its performance. Occasionally, when descriptions were provided, the model produced detailed explanations and correctly solved challenging problems, possibly due to random fluctuations.

Claude 3.5 demonstrated exceptional performance on Malayalam puzzles, particularly when visual aids were provided. Despite the actual Malayalam Unicode tokenization differing from our method, Claude 3.5 effectively utilized token references to achieve nearly perfect accuracy, including in grammatical tree construction. This suggests it may have leveraged pre-trained knowledge of Malayalam, associating token placeholders with known tokens. Without descriptions, Claude 3.5's grammatical applications in tree construction remained strong; however, in translation tasks, it often produced incorrect answers.

\noindent \textbf{With Unicode Encoding}
When Unicode encoding was employed, all three models demonstrated nearly perfect correctness. However, none provided a clear chain of thought explaining how the answers were derived. Notably, Gemini used the term "direct object" instead of "N-patient" as defined in the puzzle, which may indicate pre-trained knowledge of Malayalam grammar, as it did not use this term in experiments without Unicode encoding.

\subsubsection{Performance on Meroitic Puzzles}
The Meroitic puzzles involved a language pairing task and a Meroitic to Roman-script transliteration task, requiring the models to deduce two linguistic rules: Meroitic is written from right to left, and all tokens have an inherent vowel unless followed by another token.

\noindent \textbf{Without Unicode Encoding}
For the low-resource language Meroitic, all three models seldom provided chain-of-thought reasoning. GPT-4o failed to deduce critical linguistic rules, with outputs frequently incorrect and contradictory, likely due to the scarcity of training data on this rare language. The inclusion of visual aids and descriptions did not significantly enhance GPT-4o's performance; although the model occasionally mentioned visual information, its utilization was ineffective and unrelated to the answers.

Gemini struggled considerably with Meroitic puzzles, often misconstruing the task as pairing tokens with Roman script tokens, despite provided examples. In some instances, the model attempted to analyze token frequencies, but this approach was ineffective for solving the puzzles.

Claude 3.5 occasionally identified that certain token combinations, \texttt{token\_3} and \texttt{token\_6}, should be considered a single unit representing the vowel 'a', aligning with the correct linguistic rules. However, it often employed token count analysis, which proved ineffective, leading to poor results. Nonetheless, Claude 3.5's analysis demonstrated a deeper engagement with the linguistic structure of Meroitic compared to the other models.

\noindent \textbf{With Unicode Encoding}
Even with Unicode encoding, which provided correct script orientation and tokenization, GPT-4o's performance remained poor, suggesting that Unicode support alone is insufficient without adequate exposure to the language during training.

Gemini initially misunderstood the questions but showed slight improvement after additional examples were provided. However, it tended to reuse provided examples directly as answers, indicating a lack of training on Meroitic data and difficulty in generalizing to this low-resource language.

Claude 3.5 achieved the best performance among the models, correctly answering 6 out of 12 pairing questions. However, the model did not provide chain-of-thought explanations, leaving the rationale behind its improved performance unclear. Given the low prevalence of Meroitic in training data, it is unlikely that Claude 3.5 was extensively pre-trained on this language. The improved results may be attributed to the additional information provided by Unicode encoding, such as script directionality and accurate tokenization, which facilitated better problem-solving. In translation tasks, despite not constructing a comprehensive word correspondence table, Claude 3.5's transliterations from Meroitic to the Roman script exhibited the highest similarity to the correct answers.







\appendix



\end{document}